\documentclass[runningheads]{llncs}
\usepackage{graphicx}
\usepackage{verbatim} 
\begin{document}
\title{Budget Sensitive Reannotation of Noisy Relation Classification Data  Using Label Hierarchy}
\titlerunning{Budget Sensitive Reannotation}
% If the paper title is too long for the running head, you can set
% an abbreviated paper title here
%
\author{Akshay Parekh, Ashish Anand, Amit Awekar}
\authorrunning{Parekh et al.}
% First names are abbreviated in the running head.
% If there are more than two authors, 'et al.' is used.
%
\institute{Indian Institute of Technolgy Guwahti, Kamrup, Assam - 781039}

\maketitle              % typeset the header of the contribution
\begin{abstract}

Large crowd-sourced datasets are often noisy and relation classification (RC) datasets are no exception. Reannotating the entire dataset is one probable solution however it is not always viable due to time and budget constraints. This paper addresses the problem of efficient reannotation of a large noisy dataset for the RC. Our goal is to catch more annotation errors in the dataset while reannotating fewer instances. Existing work on RC dataset reannotation lacks the flexibility about how much data to reannotate. We introduce the concept of a reannotation budget to overcome this limitation. The immediate follow-up problem is: Given a specific reannotation budget, which subset of the data should we reannotate? To address this problem, we present two strategies to selectively reannotate RC datasets. Our strategies utilize the taxonomic hierarchy of relation labels. The intuition of our work is to rely on the graph distance between actual and predicted relation labels in the label hierarchy graph. We evaluate our reannotation strategies on the well-known TACRED dataset. We design our experiments to answer three specific research questions. First, does our strategy select novel candidates for reannotation? Second, for a given reannotation budget is our reannotation strategy more efficient at catching annotation errors? Third, what is the impact of data reannotation on RC model performance measurement? Experimental results show that our both reannotation strategies are novel and efficient. Our analysis indicates that the current reported performance of RC models on noisy TACRED data is inflated.

% This paper addresses the problem of efficient reannotation of large noisy dataset for the Relation Classification (RC) task. Our goal is to catch more annotation errors in the dataset while reannotating fewer datapoints. Existing works on dataset reannotation for the RC task have two main bottlenecks. First, they  fail to provide flexibility about how much data to reannotate. Second, they ignore the taxonomic hierarchy that exists between the relation labels. To overcome the first bottleneck, we introduce the concept of reannotation budget, allowing a user to reannotate only a subset of data. To overcome the second bottleneck, we utilize the taxonomic hierarchy of relation labels while selecting the datapoints for reannotation. The intuition of our work is to rely on the graph distance between actual and predicted relation labels in the taxonomic hierarchy.  We evaluate our reannotation strategies on the well-known TACRED dataset. We design our experiments to answer three specific research questions. First, does our strategy select novel candidates for reannotation? Second, for a given reannotation budget is our reannotation strategy more efficient at catching annotation errors? Third, what is the impact of data reannotation on RC model performance measurement? Experimental results show that as compared to the existing approaches, our reannotation strategies are novel and efficient. Our analysis indicates that the current reported performance of RC models on noisy TACRED data is inflated.

\end{abstract}
\section{Introduction}
\begin{figure}[bt]
    \centering
    \includegraphics[width=0.9\textwidth, height=4.5cm]{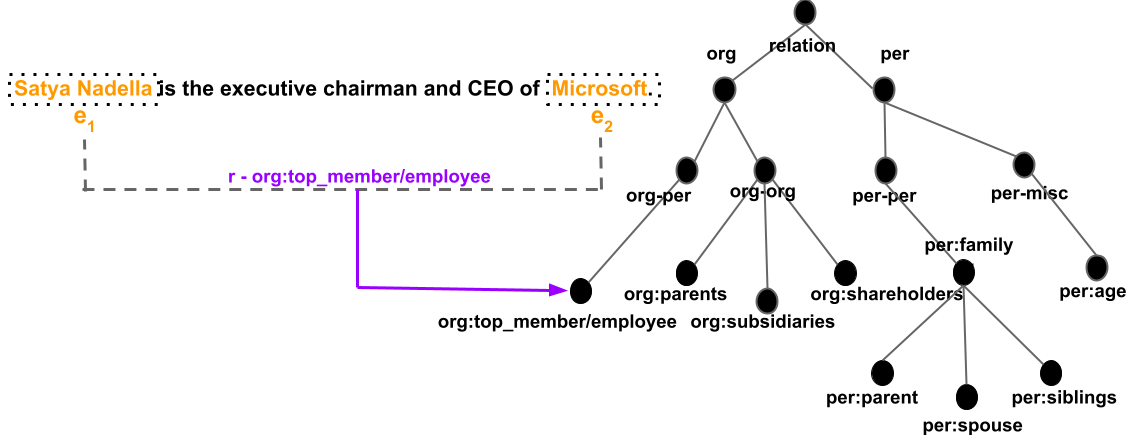}
    \caption{A sampled example from TACRED (top) and a part of a taxonomic-hierarchy for TACRED relation labels.}
    \label{fig:Example}
\end{figure}

Relation Classification (RC) is the task of predicting a relation label between a pair of real-world entities in a given natural language sentence. For example in Figure \ref{fig:Example}, the sentence has subject entity \textit{Satya Nadella} and object entity \textit{Microsoft}. Their types are \texttt{PERSON} and \texttt{ORGANIZATION} respectively. The assigned relation label for this sentence is $org:top\_member/employee$. In general, the RC task considers a five-tuple $I=<S,e_1,e_2,T_1,T_2>$ as input where $S$ is a sentence in a natural language, and $e_1$, $e_2$ are a pair of entities present in $S$. Entity $e_1$ is the subject and entity $e_2$ is the object. The type labels for these entities are $T_1$ and $T_2$ respectively. The task of Relation Classification (RC) is to assign a relation label $r$ to $I$ where $r\in R$ (the set of relation labels). Relation classification is an important NLP task with potential application in multiple domains such as knowledge graph completion \cite{kgcompletion}, question answering \cite{kbqa}, and reasoning \cite{kagnet}.

{}

The performance of state-of-the-art (SOTA) RC models is throttled in the range of 70-80\% accuracy only \cite{tacred,cgcn,spanbert,luke,improvedRE}. Various paradigms such as feature based methods, kernel based methods, and Deep Learning have been explored for designing supervised models for the RC task. All the SOTA RC models are based on Deep Learning. Performance of such models depends significantly on the quality of the dataset used for training and testing. Recent studies have shown that quality of dataset is a major bottleneck in improving the performance of supervised machine learning models for RC \cite{tacrev,retacred}.

One possible solution for improving the data quality is to reannotate the data. Existing methods for RC dataset reannotation follow two extremes. They either select only a tiny fraction for reannotation \cite{tacrev} or go for reannotation of the complete data \cite{retacred}. The existing methods also ignore the taxonomic hierarchy that exists between the relation labels.

Our work overcomes both these limitations with the following specific research contributions
\begin{itemize}
    \item{We introduce the concept of reannotation budget to provide flexibility about what fraction of dataset to reannotate.}
    \item{This is the first work that uses relation label hierarchy while selecting datapoints for reannotation for the RC task.}
    \item{We perform extensive experiments using the popular RC dataset TACRED. We show that our reannotation strategies are novel and more efficient when compared with the existing approaches.}
    \item{Our experiments suggest that the reported performance of existing RC models on the noisy dataset is inflated. The F1 score of these models drops from the range of 60\%-70\% to as low as below 50\% when tested on clean test data generated using our reannotation strategy.}
    
\end{itemize}

For the complete reproducibility, all data, models, relation hierarchy, and code from our work will be made available publicly on the Web\footnote{Link eliminated for double-blind review.} Our current experimental results are specific to the RC task. However, our reannotation strategy can be applied to any task and dataset where the label hierarchy is available. Rest of the paper is organized as follows. We review the related work in Section \ref{sec:relatedwork}. Our reannotation approach along with experimental results is presented in Section \ref{sec:ourwork}. Limitations of our work are discussed in Section \ref{sec:limitations}. Finally, we conclude and point the future work in Section \ref{sec:conclusion}.

\section{Related Work}
\label{sec:relatedwork}
Some of the earliest methods for the RC task were based on pattern extraction. They  \cite{huffman1995,patty} use syntactic patterns to construct rule for extracting relational facts from texts. To reduce the human effort in identifying relation facts, statistical relational relation extraction has been extensively explored in two directions, namely, \textit{feature-based methods} \cite{kambhatla2004,zhou2005,nguyen2007} and \textit{kernel-based methods} \cite{culotta2004,bunescu2005,bunescu2006,wang2008}. Following the success of embedding models in other NLP challenges, \cite{weston2013,gormley2015} have made efforts in utilising low-dimensional textual embedding for relational learning.

Neural Relation Extraction models introduce neural networks for capturing relational facts within text, such as \textit{recursive neural networks} \cite{recursive1,recursuve2}, \textit{convolutional neural networks (CNN)} \cite{cnn1,cnn2,cnn3}, \textit{recurrent neural networks (RNN)} \cite{tacred,bilstm1}, and \textit{attention-based neural networks} \cite{attention1,attention2}. Recently, \textit{Transformer} \cite{transformer} and \textit{pre-trained language models} \cite{bert} have also been explored for relation extraction \cite{tre,mtb,spanbert,knowbert,luke,improvedRE} and have achieved new state-of-the-arts performance.

Out of all the RC datasets \cite{mitchell2005ace,riedel2010modeling,semeval,tacred,hao2018fewrel}, TACRED \cite{tacred} is the largest and most widely used dataset. It contains more than 100 thousand sentences for 42 relation classes (Train: 68124 sentences, Dev: 22631 sentences, and Test: 15509 sentences). Sentences are collected from TAC knowledge base population 2009-2014 tasks \cite{tacred}. Each sentence contains a pair of real-world entities representing subject and object entities and is annotated with either one of the 41 relations or \textit{no\_relation}. Sentences are manually annotated with relation labels using the Amazon Mechanical Turk crowdsourcing platform, where each annotator was provided with sentence, subject and object entities, and a set of relations to choose from.

A recent study by Alt et. al. (referred to as TACRev) trained 49 different RC models and selected 5000 most miss-classified instances from TEST and DEV partitions for reannotation \cite{tacrev}. Their expert annotators were trained linguists. Out of 5000 instances, they ended up modifying 960 sentences from TEST and 1610 sentences from DEV. The revised dataset resulted in an average 8\% improvement in f1-score, suggesting the noisy nature of TACRED dataset. However, their work has two main bottlenecks. First, their method has a fixed set of sentences to reannotate. Second, they ignore the label hierarchy and use only model confidence to select sentences for reannotation.  We compare our strategies against TACRev by extending their method for the reannotation budget. Given a specific reannotation budget, we select top sentences from the dataset where RC models have the highest confidence.

Another recent study by Stoica et. al. (referred to as ReTACRED) reannotated the complete TACRED dataset using crowd-sourcing \cite{retacred}. For more effective crowd-sourcing, they have made slight modifications to the original TACRED relation list. They corrected 3936 and 5326 annotation errors in the TEST and DEV partitions respectively. They have also eliminated 2091, 3047, and 9659 sentences from TEST, DEV, and TRAIN set respectively for various reasons. We use the data from their reannotation experiment to simulate our reannotation strategies. Our work cannot be directly compared against ReTACRED as they simply use the brute-force method of complete reannotation. However, for a meaningful comparison with ReTACRED, we extend their approach by selecting a random set of sentences for a given reannotation budget.

Recently Parekh et. al. \cite{relation-hierarchy}, have collected more than 600 relations between \textit{Person}, \textit{Organization}, and \textit{Location} entity types from multiple knowledge bases and arranged them in a taxonomical hierarchy following \textit{is-a} relationship. They manually created the relation hierarchy but did not show any applications of the generated hierarchy. In this work, we create a similar relation hierarchy for TACRED relation labels, and further use that hierarchy for calculating distance between ground-truth and prediction.

The main takeaway points from the existing work can be summarized as follows:
\begin{itemize}
\item Deep Learning is the paradigm of SOTA RC models. Hence, we will focus only on Deep Learning based models for our work.

\item TACRED is the RC dataset used by SOTA models. Also, reannotation data for the complete TACRED dataset is availble. Hence, for our experiments we will used the TACRED dataset.

% \item Existing both the works on RC dataset reannotation are rigid. They fail to provide flexibility about how much data to reannotate. Hence, we will introduce the concept of reannotation budget to overcome this limitation.

% Akshay Edit 
\item Both the existing works on RC dataset reannotation are rigid. They fail to provide flexibility about how much data to reannotate. Hence, we will introduce the concept of reannotation budget to overcome this limitation.

\item The concept of taxonomic hierarchy of relation labels is not yet used for the RC dataset reannotation. We will build such a hierarchy of relation labels in the TACRED dataset and use it for selecting datapoints for reannotation.
\end{itemize}

\section{Our Work}
\label{sec:ourwork}
Considering the insights from related work, our goal is to build a budget sensitive reannotation method and show its effect on TACRED dataset and recent deep learning based models. Consider an RC dataset $D$. Let $N$ be the sent of noisy or mislabelled sentences in $D$. As $N$ is a subset of $D$, $\vert N \vert \leq \vert D \vert$. Let $R$ represent the set of sentences that we are going to reannotate from $D$. Now, our reannotation budget is $\vert R \vert$. In other words, we can afford to reannotate only $\vert R \vert$ sentences out of $\vert D \vert$. The goal of any reannotation method should be to maximize the $\vert N \cap R \vert$. In the ideal case, $N$ and $R$ should be the identical sets.

\begin{figure*}[bt]
    \centering
    \includegraphics[width=1.0\textwidth, height=3.4cm]{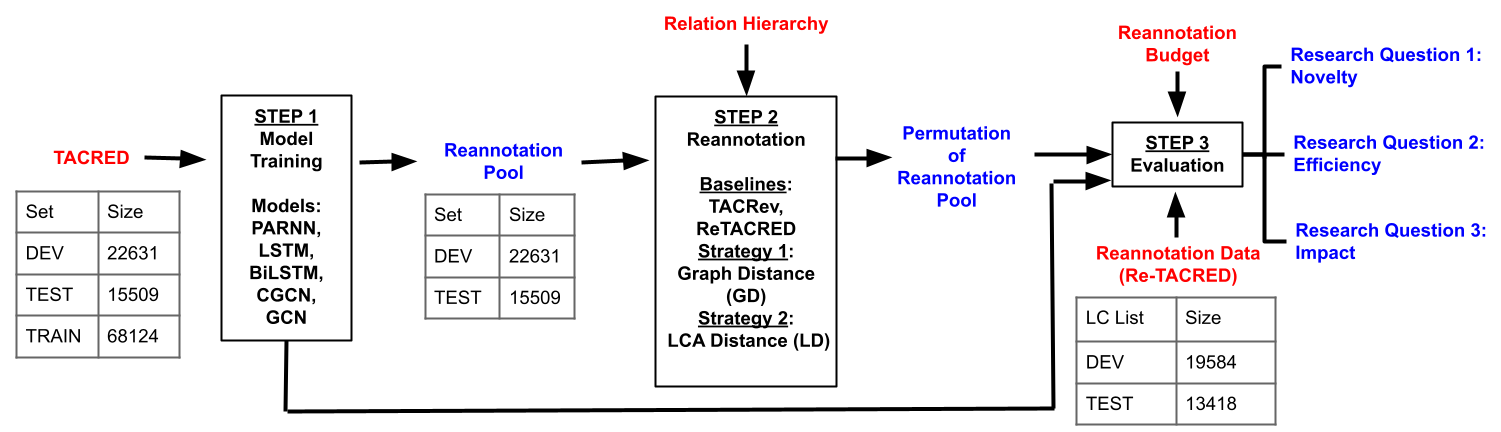}
    \caption{Overview of our work}
    \label{fig:Workflow}
\end{figure*}

Given a reannotation budget $\vert R \vert$, the immediate follow-up problem is: what subset of the data should we reannotate? If our reannotation budget permits us to reannotate a significant part of the data ($\vert R \vert \approx \vert D \vert$) then the strategy for selecting data points for reannotation does not matter. However, in the real world, the reannotation budget is far small as compared to the dataset size ($\vert R \vert \ll \vert D \vert$) for two reasons. First, reannotation is a costly and time-consuming task. Second, redundant efforts in reannotating a correctly labeled data point should be avoided. In the context of the RC task, this paper proposes two strategies for selecting data points for reannotation. Our approach capitalizes on the taxonomic hierarchy of relation labels which is largely an ignored aspect of the RC datasets. For each data point, we compute the graph distance between the actual label provided in the dataset and the predicted label using an ensemble of RC models. Data points with a higher value for this distance are given higher priority for the reannotation task. The overview of our work is presented in Figure~\ref{fig:Workflow}. Our work is divided into three steps: Model Training, Reannotation, and Evaluation. These steps are described in the following subsections.

{}

\subsection{Model Training}\label{modelTraining}
The first step trains multiple RC models using the original TACRED dataset. However, we transform the relation labels to match the ReTACRED labels. This transformation is necessary as we have reannotation data available from ReTACRED only. For our experiments we have selected following five Deep Learning based RC models: PA-RNN, LSTM, and Bi-LSTM following \cite{tacred}, CGCN, and GCN \cite{cgcn}. We have trained these models using the same set of hyper-parameters as mentioned in their original papers. This step can be further improved by training more RC models. However, we were limited by available computing infrastructure.

All the input word vectors are initialised using pre-trained 300-dimensional Glove vectors \cite{glove} for CGCN and GCN and 200-dimensional Glove vectors for LSTM, BiLSTM, and PA-LSTM. For training GCN and CGCN, we used the hyper-parameters following \cite{cgcn}. We used LSTM hidden size and feedforward hidden size as 200. We used two GCN layers and two feedforward layers, SGD as optimizer, initial learning rate of 1.0 which is reduced by a factor of 0.9 after epoch five. We trained the model for 100 epochs. We used word dropout of 0.04 and  dropout of 0.5 to LSTM layers. All other embedding (such as NER, POS) size is fixed as 30. For training PALSTM, LSTM, and BiLSTM, we followed \cite{tacred} for hyper-parameters. We have used 2 layer stacked LSTM layers for all the models with hidden size of 200. We used \textit{AdaGrad} with learning rate of 1.0 which is reduced by a factor 0.9 after 20th epoch. We have trained the model for 30 epochs. We used word dropout of 0.04 and  dropout of 0.5 to LSTM layers. All other embedding (such as NER, POS) size is fixed as 30.

\subsection{Reannotation}
The second step simulates the reannotation process by creating a specific permutation of the data points in the reannotation pool. For our experiments, we have considered TEST and DEV partitions of the TACRED dataset as the reannotation pool. The reannotation data is not available for the TRAIN partition. In our simulation, the crowd-sourced workers will reannotate only a subset of data points from the reannotation pool depending on the reannotation budget. We have experimented with four reannotation strategies: two from the literature (TACRev\cite{tacrev} and ReTACRED\cite{retacred}) and our proposed two strategies. Given a particular budget for reannotation, we have to select a set of sentences from the reannotation pool to perform the reannotation task.  A reannotation strategy creates a ranked list of the reannotation pool. The goal of a reannotation strategy is to ensure that sentences having annotation errors should appear at the top of the ranked list. Such ranking is expected to use a given reannotation budget effectively. Reannotating all the sentences in the reannotation pool will require a large amount of resources. In the real world, we can typically afford to reannotate only a fraction of the reannotation pool. We expect that a good reannotation strategy should be able to catch most of the annotation errors with a relatively small reannotation budget. 

A trained RC model when presented with a data point for relation classification, assigns a relation label along with the confidence score. This score indicates how confident the model is while assigning the relation label. The TACRev strategy uses multiple RC models for ranking the data points based on the confidence of the RC models. In our experiments, we have used five RC models mentioned in subsection ~\ref{modelTraining}. The ReTACRED method does not have the concept of ranking data points. It is a brute-force strategy that simply reannotates the whole reannotation pool. To simulate the ReTACRED strategy with the reannotation budget, we simply pick up random sentences from the reannotation pool to perform the reannotation task.

Our reannotation strategy is based on the taxonomic-hierarchy prepared by Parekh et al. \cite{relation-hierarchy}. Following their work, we arranged all relations from the TACRED dataset in a taxonomic hierarchy. The relation labels are arranged as a tree. Each relation label has a parent and multiple or zero children. Please refer to Figure \ref{fig:Example} for an excerpt of the relation hierarchy. With such a hierarchy, it becomes feasible to measure the error in the model prediction. The complete taxonomic hierarchy is available for download along with the code. Both our reannotation strategies utilize this hierarchy for selecting candidates for reannotation.

Consider a sentence $S$ with its relation label annotated in the dataset as $AR(S)$. Here $AR$ stands for Actual Relation. We have a list of $K$ different RC models: $M_1$ to $M_K$. Each model $M_i$ assigns relation label $PR_i(S)$ to the sentence $S$. Here $PR$ stands for Predicted Relation. Both $AR(S)$ and $PR_i(S)$ can be located as nodes in the relation hierarchy tree. Our first reannotation strategy measures the length of the path between these two nodes in the tree. We refer to this strategy as the Graph Distance (GD) strategy. For each sentence $S$, we compute the following score:

$GD(S)= (\sum_{i=1}^{K} f_i(S))/K$,where

$f_i(S)=Distance(AR(S),PR_i(S))$

The $Distance$ function computes the length of shortest path between given two tree nodes. This is a basic problem in graph algorithms and can be solved in time complexity $O(h)$, where $h$ is the height of the tree. Please refer to Figure~\ref{fig:Example}. Let us assume that for a given sentence $S$, $AR(S)$ is $per:parent$ and $PR(S)$ is $per:age$. In this case, the shortest path between these two labels is ($per:parent$, $per:family$, $per-per$, $per$, $per-misc$, $per:age$) with the $Distance$ as five. The function $f_i(S)$ measures the disagreement between the prediction of model $M_i$ and the label given in the dataset. High value for $f_i(s)$ indicates high disagreement between model $M_i$ and the currently assigned label in the dataset. The $GD(S)$ score computes the average across $K$ models. Our Graph Distance reannotation strategy considers sentences with a high $GD$ score as preferred candidates for reannotation. Intuitively, we are selecting sentences where multiple RC models have strong disagreement with the currently assigned label.

Our second strategy fine-tunes the computation of disagreement between model prediction and currently assigned label in the dataset. We locate both $AR(S)$ and $PR_i(S)$ in the relation hierarchy and compute their Lowest Common Ancestor (LCA). The $LCA$ computation in the tree is also a basic graph problem and can be solved in time complexity $O(h)$, where $h$ is the height of the tree. Let us consider the same example with $AR(s)$ as $per:parent$ and $PR(S)$ as $per:age$. In this case, the $LCA$ of these two labels is the node $per$. Consider the path from the root of the relation hierarchy to the node $AR(S)$. The LCA node is the point at which the model prediction starts differing from the currently assigned label in the dataset. We compute the following score for each sentence $S$:

$LD(S)= (\sum_{i=1}^{K} g_i(S))/K$,where

$g_i(S)=Distance(AR(S),LCA(AR(S), PR_i(S)))$

The function $g_i(S)$ is an improvement over the previously defined function $f_i(S)$. Instead of considering the complete path between $AR(S)$ and $PR_i(S)$, we are considering only the part up to the LCA of these two nodes. We call this reannotation strategy as LCA Distance (LD) strategy. Each reannotation strategy creates a permutation of the reannotation pool by sorting sentences in the descending order of the corresponding scores.

\subsection{Evaluation}
After Step 2, each reannotation strategy will create its own permutation of the reannotation pool. Based on the reannotation budget $\vert R \vert$, we will select the top $\vert R \vert$ data points from each permutation. While comparing our strategies with the TACRev and ReTACRED, we need to answer two important questions. First about novelty: do our strategies provide novel candidates for reannotation? Second about efficiency: do our strategies catch more annotation errors for a given reannotation budget?

\begin{figure}[bt]
    \centering
    \includegraphics[width=\textwidth]{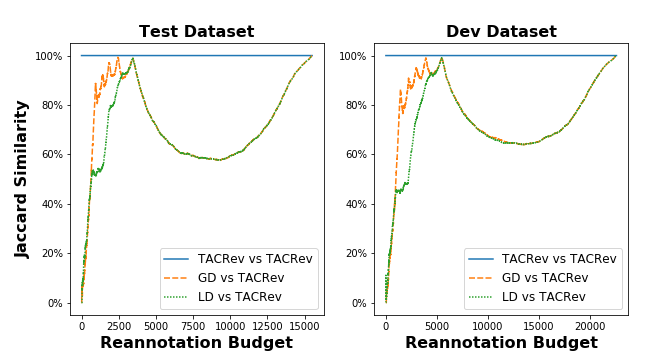}
    \caption{Similarity of a reannotation strategy with TACRev.}
    \label{fig:intersection}
\end{figure}{}

\subsubsection{Novelty in Candidate Selection}
 Novelty of GD and LD against ReTACRED is trivial because ReTACRED selects random data points for reannotation. Please refer to Figure \ref{fig:intersection} for the analysis of novelty against TACRev. The X-axis indicates the reannotation budget. Y-axis indicates the Jaccard similarity, that is the size of the intersection divided by the size of the union. The similarity of TACRev with itself is always 100\%. We can observe that for a low reannotation budget, the similarity score grows quickly and approaches 100\%. This indicates that initially, all strategies choose similar candidates for reannotation. For medium budget value, the similarity falls significantly. This indicates that our strategies differ significantly from TACRev for medium-budget value. For a higher reannotation budget, the similarity score again approaches 100\% as expected. When the reannotation budget is high, we are going to reannotate almost all the dataset. In such a scenario, there is hardly any chance for novel candidate selection.

\subsubsection{Efficiency}

ReTACRED performed complete reannotation of TACRED data \cite{retacred}. They observed that the set of noisy datapoints ($N$) in TACRED are: 5326 sentences from DEV partition and 3936 sentences from TEST partition. To evaluate the performance of our reannotation strategies, the obvious way is to carry out the reannotation task. However, we simulated the performance of our reannotation strategies using the set $N$ for two main reasons. First, the set $N$ is carefully prepared by designing a large and multi-stage crowdsourcing experiment by ReTACRED authors. Second, we did not have the financial budget available to carry out the large-scale reannotation task. We take the labels from ReTACRED as gold standard labels.

\begin{figure}[bt]
    \centering
    \includegraphics[width=\textwidth]{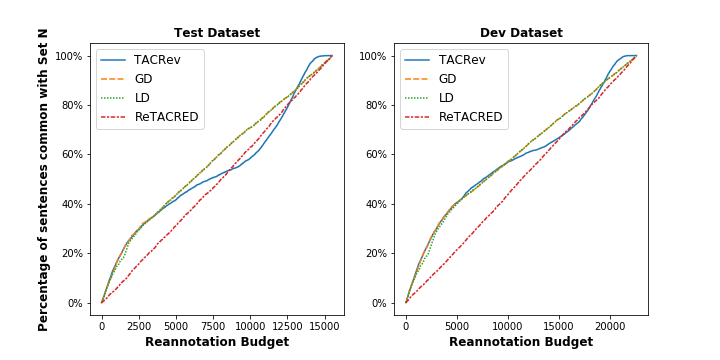}
    \caption{Annotation errors corrected by various reannotation strategies.}
    \label{fig:common}
\end{figure}

Please refer to Figure \ref{fig:common}. The X-axis indicates the reannotation budget. Y-axis depicts the percentage of sentences from the set $N$ that we can find in the given budget using a particular reannotation strategy. For example, when we reannotate the top 1000 sentences from the TEST partition using the LD strategy, it contains 566 sentences, around 14.4\% from the set $N$. We can observe that for a low reannotation budget, GD, LD, and TACRev have the same efficiency. This is expected as their similarity was close to 100\% for the low reannotation budget. However, for the medium reannotation budget our strategies outperform TACRev. For the very high value of the reannotation budget, TACRev performs slightly better than our strategies. The efficiency of GD and LD strategies is almost the same as their lines are overlapping in the figure. Efficiency of ReTACRED grows linearly as it chooses random sentences for reannotation.

{}

\subsubsection{Impact on Model Performance Measurement}
\begin{figure}[bt]
    \centering
    \includegraphics[width=\textwidth]{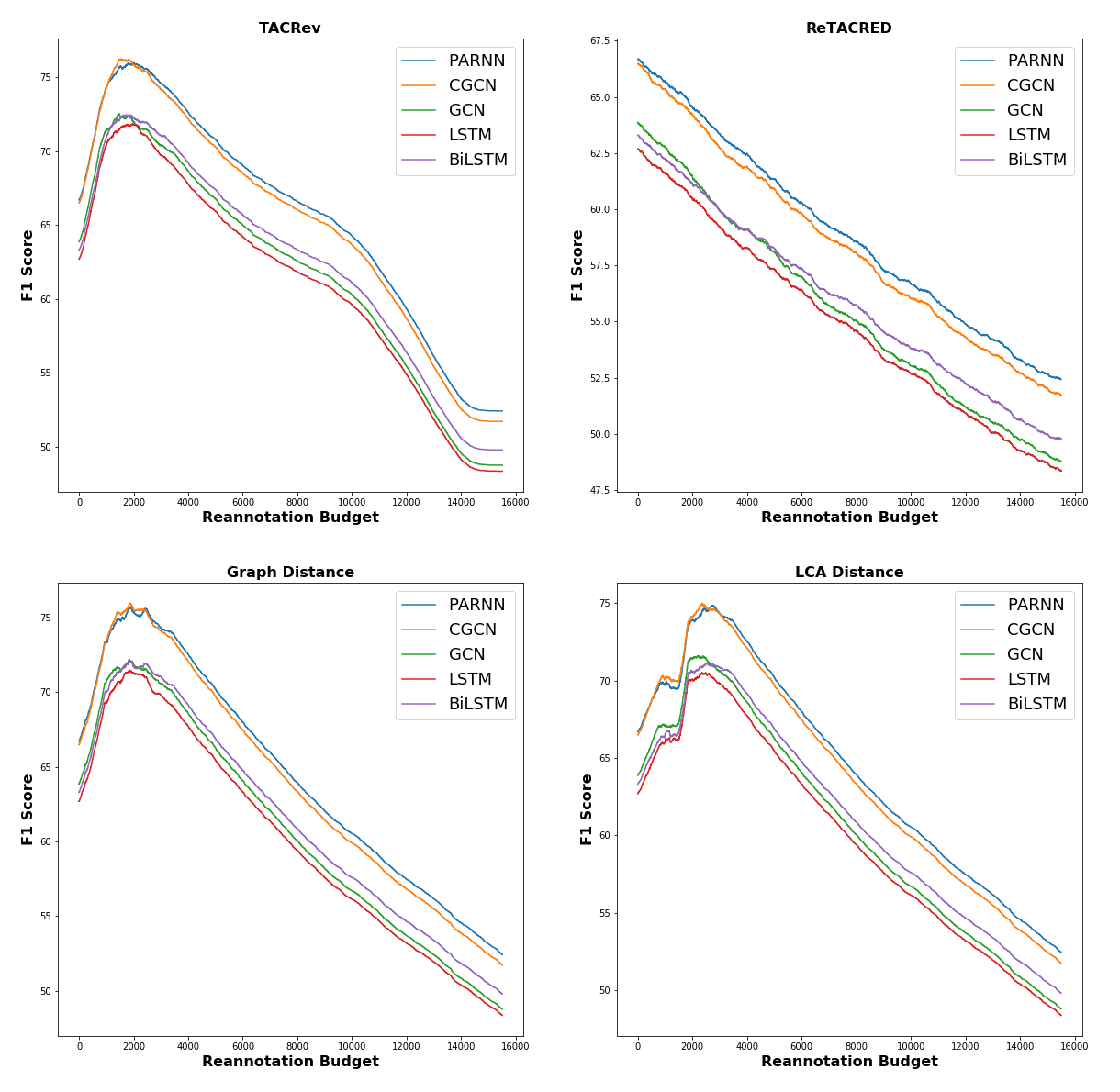}
    \caption{Performance of RC models at different Reannotation Budget.}
    \label{fig:match}
\end{figure}

Please refer to Figure \ref{fig:match}. The X-axis indicates the reannotation budget. Y-axis indicates the F1 score of each model. Please note that we are only reannotating the TEST data here. The models are trained on the noisy TACRED dataset. We can observe that for three reannotation strategies (TACRev, GD, and LD), initially, the model performance improves with reannotation. This indicates that RC models agree with the initial reannotations. However, the performance of all models falls significantly with a higher reannotation budget. The test dataset is the cleanest when the reannotation budget is maximum. The least value of the F1 score for all models at this point indicates that the original performance reported in the literature for these models was inflated. The main reason for this inflated value is the significant noise in the TRAIN partition of the TACRED dataset.

{}

None of the previous works could catch this trend for the following reason. We have conducted experiments here with noisy training data and a varying degree of noise reduction in the test data. However previous works can be classified in the following categories

   \textbf{ Noisy training and noisy test data:} Various works that directly used the TACRED data are training their RC models on noisy TRAIN partition of TACRED \cite{cgcn}. They are also testing their RC models on noisy TEST data. As a result, their reported F1 score is in the range of 60\% to 70\%. This corresponds to reannotation budget zero in Figure \ref{fig:match}.
    
    \textbf{Noisy training and partially clean test data:} The TACRev study had reported an average 8\% improvement in the F1 score with the reannotation budget of about 2300 sentences from the TEST partition \cite{tacrev}. Our results for TACRev in Figure~\ref{fig:match} correlate with their reported numbers. We can observe that while using the TACRev reannotation strategy, performance of all models peaks with the reannotation budget of around 2300. However, their results are a special case of our experiments with a fixed reannotation budget of 2300. We can observe a similar trend for our reannotation strategies GD and LD in Figure \ref{fig:match}. Further cleaning of TEST dataset beyond the budget of 2300, actually results in the performance drop for all RC models and all reannotation strategies.
    
    \textbf{Partially clean training and completely clean test data:} ReTACRED performed complete reannotation of the DEV and TEST partitions of the TACRED dataset \cite{retacred}. They did not perform any reannotation of the TRAIN partition. However, they eliminated around ten thousand potentially noisy sentences from the TRAIN partition. As a result, they report at least 10\% improvement in F1 score for various RC models as compared to previous works. This case cannot be captured in Figure \ref{fig:match} because we are not making any changes to the TRAIN partition of the TACRED dataset.
    
    We have extended the ReTACRED approach as a reannotation strategy while keeping the training data unchanged and progressively cleaning the test data depending on the reannotation budget. In Figure \ref{fig:match}, we can observe that for ReTACRED strategy, the model performance goes on monotonically decreasing with the reannotation budget. This is expected as the order of the sentences for reannotation is random. From other three reannotation strategies we can observe that reannotation of only about top 2500 sentences helps RC models to boost their performance. All these 2500 sentences are ordered randomly in the ReTACRED strategy. Reannotation of the rest of the sentences from the TEST partition does not agree with RC model prediction and brings down the model performance.

\section{Limitations of Our Work}
\label{sec:limitations}
The first limitation of our work is that we are trusting the annotations from ReTACRED as the ground truth. In reality, the ReTACRED annotations are also done using crowd-sourcing. However, as compared to the original TACRED dataset, they have taken more precautions to avoid annotations errors. For example, they have rearranged the relation labels to avoid confusion in labeling. They have also created better instructions for the crowd-source workers. One way to further reduce the possibility of errors in the data is to get annotations from domain experts. However, considering the scale of dataset, it is a challenging task.

The second limitation of our work is that we have not yet dealt with reannotation of TRAIN partition. It will be interesting to see if the concept of reannotation budget helps to significantly improve the quality of training data while annotating only a small fraction of the training data. Currently, we are working on an automatic method for reannotating the training data using an ensemble of RC models.

The third limitation of our work is that it requires a taxonomic hierarchy of relation labels. We have created the relation hierarchy manually for the TACRED dataset. It will be helpful if such a hierarchy can be created automatically from the RC dataset.

\begin{comment}
\subsection{Why to use this Reannotation scheme?}
It has been observed across various domains that crowd annotated dataset are often noisy as humans are tend to make errors. Annotating dataset employing human (or expert) is itself time and cost intensive process. Reannotating the entire dataset following the same will further increase the complexity. Our proposed method has following advantages (i) flexibility in terms of number of instances to reannotate depending on the budget (ii) cost-effective, as it is based on models' prediction and label hierarchy.

\end{comment}

\section{Conclusion and Future Work}
\label{sec:conclusion}
We presented two reannotation strategies in this paper: GD and LD. We also introduced the concept of reannotation budget to provide flexibility about how much data to reannotate. As compared to existing strategies, our strategies are novel and efficient. These reannotation strategies also help us understand the true performance of RC models trained over the noisy TACRED data. We plan to extend this work by designing strategies for selective reannotation of training data and evaluating its impact on model training. 

% The most important conclusion from this work is that we should not ignore the taxonomic hierarchy between the labels. It is an important but underutilized aspect of the data. A high rate of reannotation indicates that crowd-sourced labels are not trustworthy, especially for complex NLP tasks such as Relation Classification. We should not think of data annotation as a one-time task. Rather it should be treated as a budget-sensitive and iterative process where data is labelled multiple times.

The important conclusions from this work are: (i) relation between labels is an important but underutilized aspect of the data. It can helpful in numerous ways to improve dataset. (ii) A high rate of reannotation indicates that crowd-sourced labels are not trustworthy, especially for complex NLP tasks such as Relation Classification. Hence, data annotation should not be considered as a one-time task. Rather it should be treated as a budget-sensitive and iterative process where data is labelled in multiple iterations.

%
% ---- Bibliography ----
%
% BibTeX users should specify bibliography style 'splncs04'.
% References will then be sorted and formatted in the correct style.
%
\bibliographystyle{splncs04}
\bibliography{sample-base.bib}
%
% \begin{thebibliography}{8}
% \bibitem{ref_article1}
% Author, F.: Article title. Journal \textbf{2}(5), 99--110 (2016)

% \bibitem{ref_lncs1}
% Author, F., Author, S.: Title of a proceedings paper. In: Editor,
% F., Editor, S. (eds.) CONFERENCE 2016, LNCS, vol. 9999, pp. 1--13.
% Springer, Heidelberg (2016). \doi{10.10007/1234567890}

% \bibitem{ref_book1}
% Author, F., Author, S., Author, T.: Book title. 2nd edn. Publisher,
% Location (1999)

% \bibitem{ref_proc1}
% Author, A.-B.: Contribution title. In: 9th International Proceedings
% on Proceedings, pp. 1--2. Publisher, Location (2010)

% \bibitem{ref_url1}
% LNCS Homepage, \url{http://www.springer.com/lncs}. Last accessed 4
% Oct 2017
% \end{thebibliography}
\end{document}